# DECISION MAKING "BIASES" AND SUPPORT FOR ASSUMPTION-BASED HIGHER-ORDER REASONING[1]

By
Marvin S. Cohen

Decision Science Consortium, Inc.
1895 Preston White Drive, Suite 300
Reston, VA 22091
(703)620-0660

## ABSTRACT

Unaided human decision making appears to systematically violate consistency constraints imposed by normative theories; these biases in turn appear to justify the application of formal decision-analytic models. It is argued that both claims are wrong. In particular, we will argue that the "confirmation bias" is premised on an overly narrow view of how conflicting evidence is and *ought* to be handled. Effective decision aiding should focus on supporting the contral processes by means of which knowledge is extended into novel situations and in which assumptions are adopted, utilized, and revised. The Non-Monotonic Probabilist represents initial work toward such an aid.

## Conflicting Evidence

Two broadly different approaches to conflicting evidence have been adopted in systems that perform inference under uncertainty. In logic-based systems that are capable of "non-monotonic" reasoning (e.g., McDermott and Doyle, 1980), if it is possible to derive a contradiction from a set of statements, then one or more of those statements must be false. Suppose, for example, we start from the following beliefs:

Argument #1: $p, p \rightarrow q$

Argument #2: $r, r \rightarrow \neg q$

From these two arguments, we could infer an impossibility: the truth of both $q$ and $\neg q$. To remove the inconsistency, at least one of the beliefs responsible for it must be revised. We *know* we are wrong about $p$, $p \rightarrow q$, $r$, or $r \rightarrow \neg q$

A quite different approach has been adopted in systems that quantify and combine *degrees* of belief, like probability theory, fuzzy logic, or Shafer-Dempster theory. Suppose we believed:

Argument #1: $S(p) = .99; S(p \rightarrow q) = .99$.

Argument #2: $S(r) = .99; S(r \rightarrow \neg q) = .99$.

where $S(\cdot)$ denotes probability, Shaferian belief, or any comparable measure. Although it may follow that we have very strong evidence for $q$ and very strong evidence for $\neg q$, there is neither logical contradiction nor probabilistic incoherence. Even strong evidence may be imperfectly associated with hypotheses. Legitimate evidential arguments may, therefore, point in different directions as long as each argument falls short of conclusive proof. Thus, it is *conceivable* that all our original beliefs were correct. The question remains, nevertheless, whether it is *plausible*: Does the mere addition of a graded support measure (which may after all approach arbitrarily close to 1.0) absolve us of the responsibility to re-examine our beliefs? More pragmatically, must we be satisfied with equal support for $q$ and $\neg q$?

The first approach to conflicting evidence is *epistemic*: conflict is regarded as a symptom of faulty beliefs and is used as an opportunity to correct them--by explicitly identifying potentially erroneous steps in the conflicting arguments. The second approach may be loosely referred to as *stochastic*: conflict among imperfect arguments is *expected* to occur by chance, and it is dealt with not by changing the arguments, but by statistically aggregating them.

---

[1] This research was supported in part by Contract No. MDA903-86-C-0383 from the Army Research Institute. The COTR is Dr. George Lawton.

Each approach has virtues: On the one hand, the "stochastic" view, unlike the epistemic, permits gradations of belief; moreover, belief revision in epistemic systems is often arbitrary since there is no principled way to select one culprit from among the many beliefs responsible for a contradiction. On the other hand, the stochastic approach is likely to "resolve" conflict in ways that are unconvincing and that fail to extract permanent lessons that might improve future inferences. Resolutions of conflict by stochastic methods are typically either too bland (as above) or too definitive, as in the following cases:

- Argument #1 strongly supports hypothesis $S_1$, allows a very small chance that $S_2$ is correct, and provides no support for $S_3$; Argument #2 strongly supports hypothesis $S_3$, allows a very small chance that $S_2$ is correct, and provides no support for $S_1$. Statistical aggregation (Bayes' Rule, Dempster's Rule, etc.) results in 100% belief in $S_2$, which both sources regarded as highly unlikely (cf., Zadeh, 1984; Cohen, 1986).

- According to Argument #1, $\neg S$ is impossible; according to Argument #2, $\neg S$ is favored $10^{10}$ to 1. The result: 100% belief in S.

For most people, these conclusions will seem premature. Even when conflicting arguments have been expressed numerically, people look for *reasons* for the conflict.

An implication of the epistemic approach is that beliefs are often subject to retraction, i.e., they are assumptions. In the next sections, we will argue that the so-called "confirmation bias" is not well characterized as a violation of numerical coherence constraints. Rather, it is a result of efforts to utilize highly functional knowledge structures in contexts where the probabilities and utilities that would be required by decision analysis are not known with confidence. As a result, traditional decision aids may simply miss the point. Gaps in knowledge are often filled by a process of reasoning in which assumptions are implicitly or explicitly adopted, utilized, and revised. Decision aids should support, rather than ignore, reasoning at this level.

## Confirmation Bias in Interpreting Evidence

In both the laboratory and the real world, people seem to persevere in beliefs despite evidence that appears on normative grounds to be sufficient to disconfirm them. Lord, Ross, and Lepper (1979), for example, gave subjects descriptions of two studies of the impact of capital punishment on murder rate: one supportive of capital punishment, the other opposing. The order of presentation was counterbalanced, as was the use of two different statistical methodologies. Subjects who initially favored capital punishment were highly critical of the study opposing it and appreciative of the study supporting it, regardless of which methodology happened to be used in each. The reverse was true for subjects who initially opposed capital punishment. Not surprisingly, then, belief in the initial hypothesis was strengthened for both groups by the study which supported it, but not weakened by the study that opposed it. The net result of reviewing both studies, one positive and one negative, was that *both* sets of subjects became *more* confident in the correctness of their own position. At DSC, Tolcott et al. (1987) found virtually identical results in a study of Army battlefield tactical intelligence, using experienced military intelligence officers. Different initial hypotheses about the likely enemy course of action (attack in the South versus attack in the North) were induced by slight variations in the scenario settings for different groups of subjects. Regardless of a subject's initial hypothesis, confidence started out high and tended to increase in time as items of new information (both favoring and opposing the hypothesis) were received. There was a tendency to rate individual items of evidence as favoring the initial hypothesis, regardless of the item or the hypothesis. "Negative evidence," i.e., the non-occurrence of events that were expected based on the hypothesis, had only a small disconfirming effect for the most experienced subjects.

Although these findings are disturbing, the moral is less clear. For any of these subjects taken alone, coherent subjective degrees of belief could be constructed. Thus, *both* groups of subjects may have been consistent (with respect to their own beliefs about the diagnosticity of evidence) in growing more confident about their initial hypotheses. Perhaps one of the groups is *wrong* in its assessments; but that is a substantive matter, not a formal error in reasoning. The overall pattern of results, however, suggests that reactions to data were influenced by prior hypotheses. Assessments of diagnosticity, had they been made prior to (or otherwise independently of) formulating an initial opinion, might well have been inconsistent with the way subjects actually did react to the evidence. Nevertheless, clever Bayesian models can be constructed here in which assessments of evidence coherently depend on prior probabilities (e.g., by conditioning probabilities on the decision maker's thoughts or judgments, or by using second-order probabilities). But such models do not tell us what it is that does or does not make sense about "confirmation bias" behavior.

Research in cognition suggests that prior expectations do play a role in interpreting data. Expectations fill gaps and help organize experiences in perception (Bruner, 1957), recall (Bransford and Franks, 1971), everyday reasoning (Minsky, 1975), and science (Kuhn, 1962). Such expectations are sometimes attributed to knowledge structures like "schemas," "frames," and "scripts." Schemas may embody a long series of experiences in a domain, capturing stable and significant correlations (cf., Brunswick, 1955). They permit successful action under conditions of incomplete and noisy data and limited time. A number of studies (e.g., Larkin, 1981) suggest that experts actually approach problems less analytically than novices. Experts appear to find solutions by recognizing similarities to problems they have solved in the past, rather than by explicitly identifying goals and carefully evaluating alternative means to achieve them (Chase and Simon, 1983; Greeno, 1983; Klein, 1980). Moreover, experts, unlike novices, perceive similarities in terms of the fundamental laws or principles in a domain rather than in terms of superficial features (Chi et al., 1981). On the other hand, schemas may also reflect the accidental and non-representative experiences of a particular individual (Schank, 1982); they may reflect erroneous underlying causal models (Holland et al., 1986); and factors such as primacy, recency, and salience may influence their content (Nisbett and Ross, 1980). Experts may be slower to find a solution than novices when the schemas they use are inappropriate for a novel problem (Sweller *et al.*, 1983).

When is it acceptable to use a schema to interpret data, rather than using data as evidence to evaluate the schema? The answer depends, in part, on how speed (obtainable by relatively automatic schema-based processing) should be traded against accuracy (Klein, 1989; Johnson and Payne, 1984). But the answer also depends on what counts as accuracy and what counts as an "error." Suppose that people always did explicitly consider the impact of new data on a schema: under what conditions should the schema be dropped in the light of the evidence? Discussions of the "confirmation bias" often sound as though simple logic or probability theory dictated the answer: contradictory evidence should prompt rejection of a contradicted theory (cf., Popper, 1959); disconfirming evidence (i.e., data that are more likely if the theory is false than if it is true) should lower confidence in the theory and its predictions.

In fact, when logic and probability are applied to even moderately-sized schemas or theories, expectation-based reasoning of this sort in no way violates logical or probabilistic principles. Consider a schema S (e.g., enemy plan to attack) in which event A (e.g., increased logistical activity in a particular sector) leads automatically *via* S to expectations of events of type B (e.g., moving up artillery in that sector) and type C (e.g., attempted penetration by enemy troops). Assuming for now that these expectations are non-probabilistic, suppose that indicators of A occur (e.g., radar reports of trucks moving toward front-line areas), but there is no sign of B: Is the schema S now invalid or inap-

73

plicable? Is it still permissible to expect C? In reality, S and the report of A imply the observation of a B *only* in conjunction with a significant number of assumptions. Some of these assumptions concern the validity of interpreting the first observation as A (Could the radar blips represent civilian traffic? Is the apparent increase in activity a statistical accident?); others concern preconditions for the linkage between A and S (Could increased logistical activity be intended to replenish a degraded defensive unit?); others concern the linkage of S and B (Does the enemy plan to omit the initial artillery barrage for purposes of surprise? Is required artillery equipment unavailable or not in working order?); and still others will concern the conditions for a successful detection of B (Could weather, foliage, or camouflage have masked the location of artillery?). The diagram shows the relevant automatic inferences (solid lines) and some of the alternative possibilities that underlie them (dotted lines).

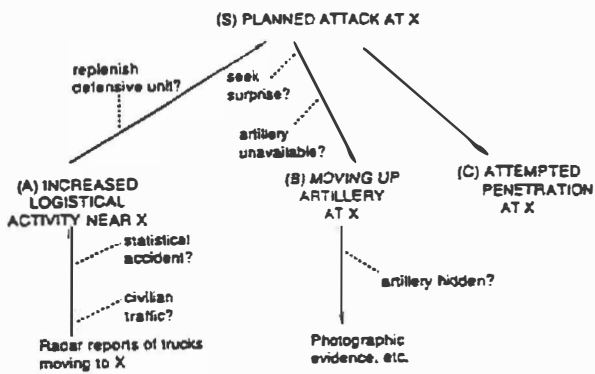

When B is not observed, therefore, logic requires only that we find a culprit somewhere in this rather large set of beliefs: we may hold onto S and reject one of the assumptions that led us (perhaps implicitly) to expect observation of B.

The "confirmation bias" is *not* a problem of logical inconsistency. Nor is it a violation of probability theory. The normative force of decision analysis is to impose axiomatic consistency on one's numerical degrees of belief as they exist at a given time. It in no way dictates what those degrees of belief ought to be or how they should change. If one's direct judgment differs from the results of an analysis, one may satisfy normative constraints either by changing the direct judgment to conform to the analysis *or* by changing various inputs to the analysis. Standard decision analytic models do not directly address the *thinking* process by means of which an analyst selects *one* consistent set of beliefs out of all those possible.

Bayes' Rule is often mistakenly regarded as a law of belief *change*, rather than a consistency constraint on simultaneously held beliefs. One problem is that beliefs regarding the hypothesis and the evidence may not have been assessed prior to receiving the evidence (Shafer, 1981; Diaconis and Zabell, 1989). But even if they have, normative rationales (e.g., avoidance of a Dutch book) permit subsequent reassessment (Horwich, 1982). Suppose we assess probabilities after observing A but before we even look for B: probability of $S = .9$; probability of not-B given $S = .2$; probability of not-B given not-$S = .8$. Then, *after* looking for and failing to observe B, the "updated" belief in S (using Bayes' Rule as if it were a law of belief change) would be: probability of S given A and not-B = $(.9)(.2)/[(.9)(.2) + (.1)(.8)] = .69$. In short, our belief in S is reduced from .9 to .69 by the nonoccurrence of B.

But we are not in fact required to keep all these assessments constant. We can react to the conflict between A and B epistemically rather than stochastically. After not observing B, we may decide our confidence in S has not been eroded very much at all; further reflection may suggest that the diagnostic importance of not observing B was less than we had thought (i.e., we have assumed artillery was available, but perhaps it is not; or we can drop the assumption that artillery would be detected; etc.). In fact, we may also choose to increase our belief in S based on A by adopting or strengthening other assumptions (e.g., that the radar blips are not civilian traffic). We might now generate an entirely new, but equally *self-consistent* set of assessments: e.g., probability of $S = .99$; probability of not-B given $S = .4$; probability of not-B given not-$S = .6$. Using Bayes' Rule with these numbers, we get: probability of S given A and not-B = $(.99)(.4)/[(.99)(.4) + (.01)(.6)] = .99$. In short, belief in S rises from .9 to .99. In this way, "negative" evidence might trigger an increase in confidence--for a perfectly coherent Bayesian decision maker.

The most plausible way to handle "confirmation bias" behavior is not to impose simple (or clever) Bayesian models of belief change, but to introduce an epistemic level of analysis. The impact of a bit of evidence is never simply given; it depends on conditions about which quite often there is little or no *direct* knowledge, and concerning which there is legitimate leeway for (implicit or explicit) assumptions. Decisions about belief involve higher-order control processes that manage assumptions while extending knowledge to novel situations.

These control processes appear to operate along a continuum of modes from "cold" to "hot" (all of them equally coherent). At the "low temperature" end of the spectrum (analogous to Piaget's "assimilation" or Thomas Kuhn's "normal science"), assumptions are adopted in order to facilitate the smooth incorporation of new data into a pre-existing framework; when data appear that seem at first blush to contradict central beliefs of the framework, the data are reinterpreted ("explained away") and the more central tenets are preserved. Note that while such "reinterpretation" of data may sometimes be virtually automatic, in other cases it may require considerable creativity. This is what scientists do every day in trying to explain unexpected new observations by means of relatively minor adjustments in existing theory. It also accounts for the perfectly reasonable practice of adjusting experimental procedures until they produce theory-predicted results (Greenwald et al., 1986). Perhaps there is, after all, a converse "disconfirmation bias" of feverishly generating alternative possibilities and never settling down to a conclusion. At the "high temperature" extreme, however (analogous to Piaget's "accommodation" and Kuhn's "revolutionary science"), this same creativity and ingenuity may be used to reject and replace even the most well-entrenched and fundamental bits of "knowledge." In this mode, one actively searches for alternative explanations for evidence that apparently *confirmed* one's previously favored beliefs! In science, this has been the mode associated with Newton, Einstein, and Darwin, but also with a rather large number of charlatans (Lysenko, Velikovsky). In sum, the evaluation of "confirmation bias" behavior is far from cut-and-dried.

## Confirmation Bias in Collecting Evidence

A "confirmation bias" appears to affect decisions about what information to collect, in addition to decisions about how to interpret it. According to Fischhoff and Beyth-Marom (1983), people seek to perform tests which, no matter what information they actually obtain, will be *interpreted* by them as confirming a particular hypothesis. Baron et al. (1988) appear to show that information collection decisions are formally incoherent. They gave subjects a description of an event, a favored hypothesis explaining the event, and questions that could be asked to determine whether or not the hypothesis was correct. Assessments for a decision analytic value of information model were obtained, providing a "normative" value of each question, determined by the chance that the answer would justify rejection of the favored hypothesis. When subjects were asked to directly rate the values of the questions, however, they consistently preferred questions that on the normative model had little or no chance of causing a change of mind. (Instead, according to Baron et al., subjects preferred questions for which there was a high probability of a positive answer given the favored hypothesis.)

We have argued that it is not necessarily irrational or unwise to adopt assumptions that construe apparently disconfirming observations as confirming. But even so, it does seem irrational to expend effort and other resources to *collect* information when one knows ahead of time how the results will be interpreted.

Baron et al. appear to believe there is only one way to interpret this apparent inconsistency: subjects overvalued questions that were undiagnostic (i.e., direct judgment of the value of questions should give way to the model). Another possibility, however, is that the model has underestimated the degree of diagnosticity of the preferred questions (i.e., model inputs should be revised in light of direct judgment). By itself, decision theory does not tell us which of these resolutions of the apparent incoherence should be adopted. Still another possibility is that standard decision analytic models are too limited to capture what is going on. Diagnosticity may be "underestimated" because subjects would (at least initially) adopt assumptions to "explain away" a disconfirming answer.

In the value of information model, the potential impact of an observation is identified with the interpretation that the decision maker assigns to it *at the present time*. If he or she would not *at present* interpret any result of a particular test as disconfirming, the test is presumed to be worthless. But the present interpretation of the evidence will almost invariably depend on assumptions, which are subject to change. Suppose, for example, that subsequent evidence continues to be interpreted as confirming the hypothesis, but that more and more work of "explaining away" is required to do this; increasingly numerous and complex assumptions are needed to "save" the hypothesis. At some point, the decision maker may realize that a much simpler overall set of beliefs and assumptions can be achieved by rejecting the favored hypothesis. (Examples of this sort of "Gestalt shift" are the scientific revolutions discussed by Kuhn, 1962.) Thus, the cumulative effect of all this "confirming" evidence is, finally, a disconfirmation! The tests that produced that evidence clearly did have value, even if no one of the tests *at the time it was performed* could have changed beliefs in the slightest degree. Actual changes in belief may not be incremental and gradual, but relatively sudden.

Here, too, clever Bayesian models could be constructed (incorporating non-independence of data among themselves, hypotheses, and prior judgments). But the most plausible account invokes an epistemic level of analysis. According to Bayesian theory, knowledge about uncertain events is fully revealed by actions, i.e., probabilities are defined by choices among gambles. This view is inadequate. A decision maker can always be forced to make a choice, but in doing so, he may well adopt assumptions that go beyond his knowledge. In a different gamble, he may adopt different assumptions, and make different choices, yet draw on exactly the same knowledge. Thus, it is necessary to distinguish the *manifest* (or behavioristic) meaning of a piece of evidence from its cognitive or *latent* meaning. The manifest meaning (e.g., the current diagnostic impact of a piece of data with respect to a particular hypothesis) is revealed by choice in a present gamble (or a present judgment), and it depends on both knowledge and a particular, possibly temporary selection of assumptions. The latent meaning refers to the total *potential* impact of the evidence on reasoning; and this can only be represented in a richer knowledge structure that includes both firmly held beliefs and the set of assumptions from which the decision maker chooses on any given occasion of choice or judgment (cf., Loui, 1986). Decision aids that assume that a belief about uncertainty is fully revealed in a single gamble or judgment and summarized in a single number forgo the opportunity to assist decision makers in the processes by which they make such choices.

### Support for Assumption-Based Reasoning

The Non-Monotonic Probabilist (NMP) is a system for inference under uncertainty in which numerical measures and an epistemic response to conflict are complementary rather than mutually exclusive (Cohen *et al.*, 1989; Cohen, 1986). Conflicting evidence is dealt with by higher-order processes that *reason about* quantitative uncertainty models; conversely, numerical measures from those models provide guidance for decisions about adopting and revising assumptions. The result is a generalization of the epistemic approach, in which belief is graded, conflict is a matter of degree rather than all-or-none, and change of belief (via assumption revision) is directed at those beliefs that are most likely to be in error.

The Self-Reconciling Evidential Database (SED) is a system for the management of evidence by intelligence analysts that incorporates the Non-Monotonic Probabilist as an inference engine. It operates in an IBM/PC computer. A more detailed discussion of SED may be found in Cohen *et al.* (1989).

SED does not assume that the meaning of an item of evidence is known precisely by the analyst from the start. A key feature of SED's approach is the phasing of argument construction to fit the natural stages of an analyst's reasoning: i.e., a "first-blush" or "normal" reaction to the evidence (which we call a "Core Position") is followed by specification of a set of possible disrupting factors. For example:

SED provides a counterbalance to the "low temperature" (or "confirmation-bias") strategy in interpreting evidence, by focusing attention on the ways in which an argument could go wrong. A powerful method for generating exception conditions is a technique that we call Conflict Resolution (Cohen, 1989; Phillips in ORD, 1982). The analyst is asked to imagine that an infallible crystal ball says the Core Position is false even though the evidence is true. Typically, the analyst will then be able to generate an explanation: e.g., the Core Position could be false even if the evidence is true, if Qualification 1 is the case. The crystal ball now tells him that the Core Position is false and the evidence is true, but Qualification 1 is also false! As a result, the analyst devises a new explanation, Qualification 2. Again, the crystal ball tells him Qualification 2 is false; and so on. The analyst is thus prompted to act as his own Devil's Advocate, exposing hidden assumptions and exploring alternative points of view.

In generating exception conditions, analysts must rely less and less on easily accessible knowledge and begin to enter "compartments" of knowledge that are not part of their ordinary "automatic" reaction to the situation. Experiments with this technique in interviews with Army intelligence officers show that it produces a rich harvest of unexpected information. It was not unusual, for example, to obtain numerous additional argument premises by means of the "crystal ball" after more direct questioning of an analyst had run completely dry. In one instance, after assessing the probability of a conclusion as 1.0, an analyst was able (by means of the crystal ball) to generate 8 different exception conditions with an average assessed probability of .31.

Numerical belief functions of arbitrary complexity are constructed in SED based on (a) a simple, qualitative specification of the separate impact of each exception condition on the argument's conclusion, and (b) degree of belief in the exception condition. Exponential growth in assessments is avoided, and non-independence among arguments *via* shared exception conditions) is automatically accommodated.

While guarding against "confirmation-bias" strategies, SED also provides an escape from "high-temperature" approaches that never settle down to a conclusion. Assumptions of some sort (e.g., about the reliability of a human source, the proper functioning of a sensor, continued accuracy of a dated observation, etc.) are necessary if definitive results are ever to be arrived at in circumstances that are even moderately novel. SED permits such assumptions to be adopted and utilized. To guard against obvious dangers, SED makes a distinction (though only a matter of degree) between assumptions and firm belief. Assumptions are (1) constrained by (though they go beyond) existing firm belief, and (2) subject to retraction if they lead to trouble--i.e., if they conflict with new evidence or with lines of reasoning supported by other assumptions.

These two safeguards correspond to two complementary ways analysts may assess their assumptions by using SED:

(1) "Bottom-up," by starting with a firm assignment of belief based on knowledge. This may be too imprecise to support an argument which the analyst presently wishes to make. Thus, the analyst may use assumptions to reallocate belief that was committed to a set of possibilities to a proper *subset* of those possibilities.

(2) "Top-down," by starting with overall belief and specifying how much of it is firm and how much the analyst would be willing to retract in case of conflict with other arguments. The analyst specifies how much of the belief in a set of possibilities he would transfer to a less precise *superset* of possibilities in case of conflict.

Typically, exception conditions are assumed false (in the absence of direct evidence) until or unless the "normal" interpretation of the evidence runs into trouble. Thus, SED permits an analyst to pursue the "automatic" or natural consequences of evidence, but prompts him or her to explicitly evaluate those consequences when they conflict among themselves. SED uses conflict as a symptom that something is wrong with one or more assumptions that led to the conflict (e.g., one or more sensors, models, human sources, etc. are not as reliable as supposed), and implements a process of higher-order reasoning that attempts to reduce conflict epistemically. Such reasoning is supported by measures both of the degree of conflict among arguments and of the degree of culpability of a given assumption for the conflict. These measures are straightforward generalizations of the logical strategy of showing a statement to be false by deriving a contradiction from it. They quantify the *chance* that a contradiction is implied, hence, the chance that one or more assumptions is in error.

In resolving conflict, the analyst may choose to drop an assumption, to collect further information, to generate new exception conditions, or even to reduce prior assessments of firm belief. SED thus focuses on an evolving, iterative understanding of the meaning or reliability of evidence, as opposed to cut-and-dried assessments of evidence strength.

SED embeds numerical uncertainty representations within a process of higher-order reasoning. Is such a higher-order process really necessary? Could the functions of epistemic conflict resolution be accomplished instead *within* a standard numerical calculus? The answer is: in principle, yes: in practice, no. To simulate the effect of epistemic conflict resolution with a numerical calculus, (e.g., Bayesian or Shaferian), a huge set of conditional assessments would be required, linking the elements of every line of reasoning to the elements of all other lines of reasoning that might possibly arise. The price of such a strategy comes not only in the sheer quantity of inputs, computational intractability, and impossibility of anticipation, but also in a loss of naturalness and modularity. Probably for these reasons, numerical inference models typically treat hypotheses about the credibility of diverse information sources or lines of reasoning as if they were independent. The result is a stochastic approach to conflict that fails to extract the real significance of conflict when it occurs. SED achieves the best of both worlds: It enables the analyst to bring to bear the conclusions of one argument on the evaluation of another without sacrificing the modularity of the different lines of reasoning. It does so by shifting responsibility for dealing with conflict from the calculus itself to higher-order processes than create and maintain the calculus.

## Conclusion

Traditional decision analytic aids require users to adopt "normative" techniques of problem solving that may differ radically from their preferred approach; such aids thus may throw out the baby (i.e., user knowledge) with the bath water (i.e., apparent user biases). In many cases, "biases" at least implicitly reflect reasonable assumption-based strategies for extending knowledge of a domain. Aids can be designed to facilitate the user's basic approach to the problem while improving performance. Such prescriptive support is a matter of helping users notice and keep track of assumptions, prompting users when their assumptions lead to trouble, and creating contexts in which existing knowledge can grow. Prescriptive support in this case focuses on control processes by means of which the user manages his or her own knowledge and reasoning.